\documentclass[sigconf]{acmart}
\AtBeginDocument{%
  }

\copyrightyear{2026}
\acmYear{2026}
\setcopyright{cc}
\setcctype{by}
\acmConference[MM '26]{Proceedings of the 34th ACM International Conference on Multimedia}{November 10--14, 2026}{Rio de Janeiro, Brazil}
\acmBooktitle{Proceedings of the 34th ACM International Conference on Multimedia (MM '26), November 10--14, 2026, Rio de Janeiro, Brazil}
\acmDOI{10.1145/3767308.3836329}
\acmISBN{979-8-4007-2213-4/2026/11}

\usepackage{multirow}
\usepackage{colortbl}
\usepackage{balance}

\definecolor{headergray}{gray}{0.95}
\definecolor{highlightgray}{gray}{0.92}
\definecolor{posgreen}{RGB}{0,150,0}

\newcommand{\gain}[2]{%
  \textcolor{posgreen}{\ensuremath{#1\,#2}}%
}

\begin{document}

\title{Scientific Graphics Program Synthesis via Dual Self-Consistency Reinforcement Learning}

\author{Juekai Lin}
\authornote{These authors contributed equally to this work.}
\affiliation{%
  \institution{Zhejiang University}
  \city{Hangzhou}
  \country{China}
}
\email{jklin7918@gmail.com}

\author{Yun Zhu}
\authornotemark[1]
\affiliation{%
  \institution{Shanghai Artificial Intelligence Laboratory}
  \city{Shanghai}
  \country{China}
}
\email{zhuyun\_dcd@zju.edu.cn}

\author{Honglin Lin}
\authornotemark[1]
\affiliation{%
  \institution{Shanghai Artificial Intelligence Laboratory}
  \city{Shanghai}
  \country{China}
}
\email{linhonglin03@gmail.com}

\author{Sijing Li}
\affiliation{%
  \institution{Zhejiang University}
  \city{Hangzhou}
  \country{China}
}
\email{suzylee@zju.edu.cn}

\author{Tianwei Lin}
\affiliation{%
  \institution{Zhejiang University}
  \city{Hangzhou}
  \country{China}
}
\email{lintw@zju.edu.cn}

\author{Liu Zheng}
\affiliation{%
  \institution{Peking University}
  \city{Beijing}
  \country{China}
}
\email{2501213330@stu.pku.edu.cn}

\author{Xiaoyang Wang}
\affiliation{%
  \institution{Shanghai Artificial Intelligence Laboratory}
  \city{Shanghai}
  \country{China}
}
\email{wangxycc@outlook.com}

\author{Wenqiao Zhang}
\correspondingauthor
\affiliation{%
  \institution{Zhejiang University}
  \city{Hangzhou}
  \country{China}
}
\email{wenqiaozhang@zju.edu.cn}

\author{Lijun Wu}
\correspondingauthor
\affiliation{%
  \institution{Shanghai Artificial Intelligence Laboratory}
  \city{Shanghai}
  \country{China}
}
\email{lijun\_wu@outlook.com}

\renewcommand{\shortauthors}{Juekai Lin et al.}

\begin{abstract}
  Graphics Program Synthesis is pivotal for interpreting and editing visual data, effectively facilitating the reverse-engineering of static visuals into editable TikZ code. While TikZ is the \textit{de facto} standard for scientific schematics due to its programmatic flexibility, its requirement for rigorous spatial precision presents a significant challenge for Multimodal Large Language Models. Progress is currently stifled by two primary gaps: (1) \emph{Data Quality Gap:} existing image-TikZ corpora often lack strict executability and reliable visual alignment; (2) \emph{Evaluation Gap:} a lack of benchmarks for both structural and visual fidelity. To address these, we present a closed-loop framework featuring: \textbf{SciTikZ-230K}, a large-scale, high-quality dataset from our Execution-Centric Data Engine covering 11 diverse scientific disciplines; \textbf{SciTikZ-Bench}, a multifaceted benchmark spanning from basic geometric constructs to intricate hierarchical schematics to evaluate both visual fidelity and structural logic. To further broaden the scope of visual-code optimization methodology, we introduce a novel \textbf{Dual Self-Consistency Reinforcement Learning} optimization paradigm, which utilizes Round-Trip Verification to penalize degenerate code and boost overall self-consistency. Empowered by these, our trained model SciTikZer-8B achieves state-of-the-art performance, consistently outperforming proprietary giants like Gemini-2.5-Pro and massive models like Qwen3-VL-235B-A22B-Instruct. Our project is available at https://github.com/JackieForest/SciTikZ.
\end{abstract}


\maketitle

\section{Introduction}
\begin{figure}[t!]
    \centering
    \includegraphics[width=1.0\linewidth]{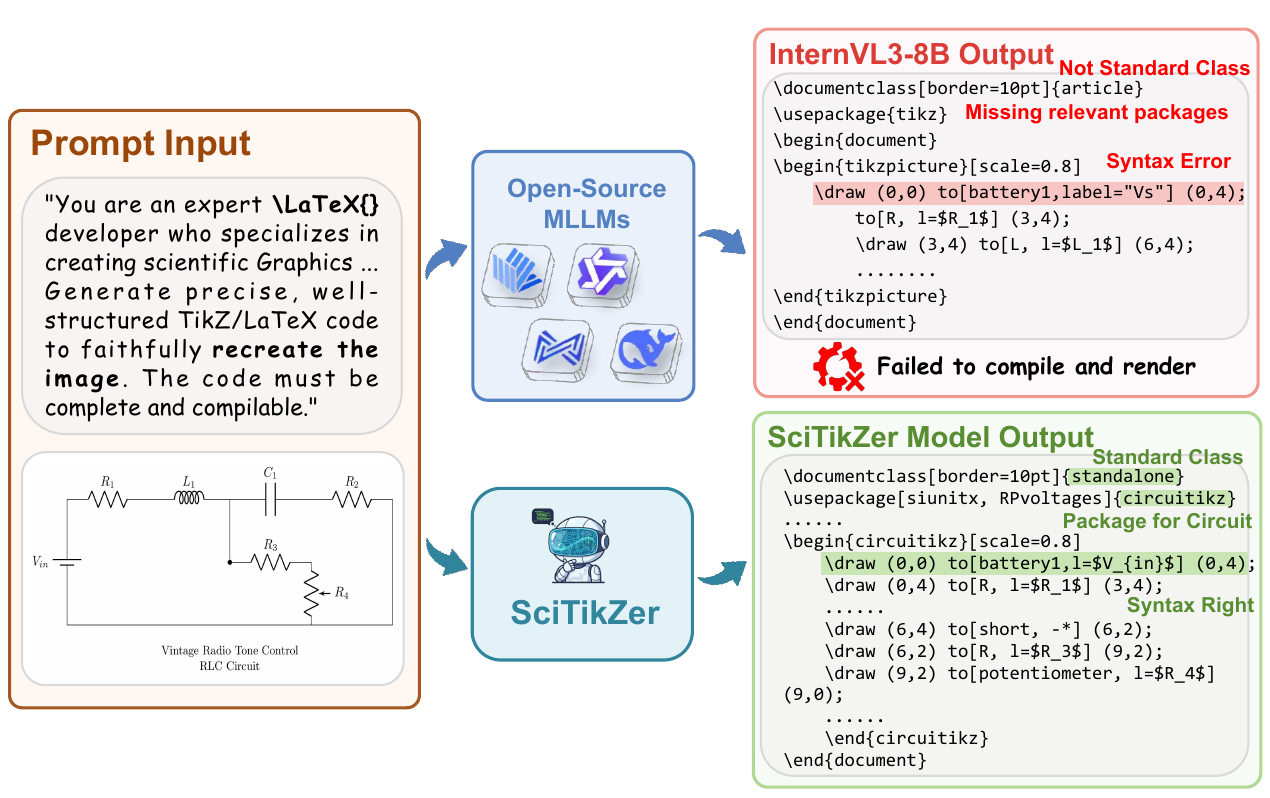}
    \caption{\textbf{Challenges in graphics program synthesis.} 
    Current open-source MLLMs struggle with the strict constraints of TikZ, exhibiting critical issues such as \textit{syntax hallucinations}, \textit{dependency omissions} and \textit{geometric misalignments}.}
    \label{fig:intro_comparison}
\end{figure}

Graphics Program Synthesis~\cite{bing2025learning, belouadi2024detikzify} enables the reverse-engineering of static raster images into editable, symbolic code, facilitating the modification and reuse of visual data. While the task spans various vector formats, the TikZ language~\cite{tantau2012graph} has established itself as the de facto standard for high-fidelity scientific schematics, such as circuit diagrams and structured flowcharts. Unlike statistical plotting libraries that tolerate automated layout approximations, TikZ demands precise coordinates, explicit symbolic primitives, and rigorous spatial definitions to represent intricate, fine-grained topological structures. This strictness renders the synthesis process highly sensitive to perturbations, where even trivial errors can trigger compilation failures or structurally degenerate artifacts during rendering and execution (Figure \ref{fig:intro_comparison}). Consequently, despite recent advances in multimodal reasoning and code generation, achieving high-fidelity schematic program synthesis still remains a formidable open problem for current Multimodal Large Language Models (MLLMs)~\cite{hurst2024gpt, comanici2025gemini, chen2024internvl, chen2024expanding, zhu2025internvl3, Qwen2-VL, li2024llava}.

Current progress in graphics program synthesis is largely driven by advances in both data scaling and methodological refinement. \textit{On the data front}, initiatives such as AutomatikZ~\cite{belouadi2023automatikz} and MathCoder~\cite{wang2025mathcoder} focus on constructing large-scale image-code pairs to facilitate effective Supervised Fine-Tuning (SFT). However, these approaches are often hampered by persistently low-quality training signals: while web-scraped data suffers from intrinsic noise, existing synthetic corpora are often low-quality, with cluttered layouts and limited structural coherence. This results in models that frequently hallucinate spatial relationships or fail to generalize beyond rigid synthetic templates in more realistic scientific settings. This issue is further compounded by \textit{a systemic evaluation deficit}. While recent benchmarks like Image2Struct~\cite{roberts2024image2struct} attempt to evaluate structured information extraction, they remain notably narrow, focusing primarily on formulas and charts rather than more complex, multi-disciplinary graphics. This lack makes it difficult to establish a standardized, comprehensive evaluation of scientific graphic synthesis. Recent \textit{methodological frameworks}, such as TikZero~\cite{belouadi2025tikzero} and DeTikZify~\cite{belouadi2024detikzify}, rely on relatively limited strategies like only CLIP-based alignment or MCTS-driven search. These paradigms primarily focus on unidirectional generation, failing to utilize TikZ’s executable nature for closed-loop training. Consequently, they struggle to balance the rigidity of SFT, which over-penalizes syntactically different yet valid variants, and the permissiveness of visual reinforcement learning (RL), which is prone to reward hacking and structural degeneracy. Ultimately, the current field remains constrained by such limited methodological strategies, lacking a generalized paradigm to ensure self-consistency between visual and symbolic representations under executable, structurally faithful program synthesis objectives.

\begin{figure}[t]
    \centering
    \includegraphics[width=1.0\linewidth]{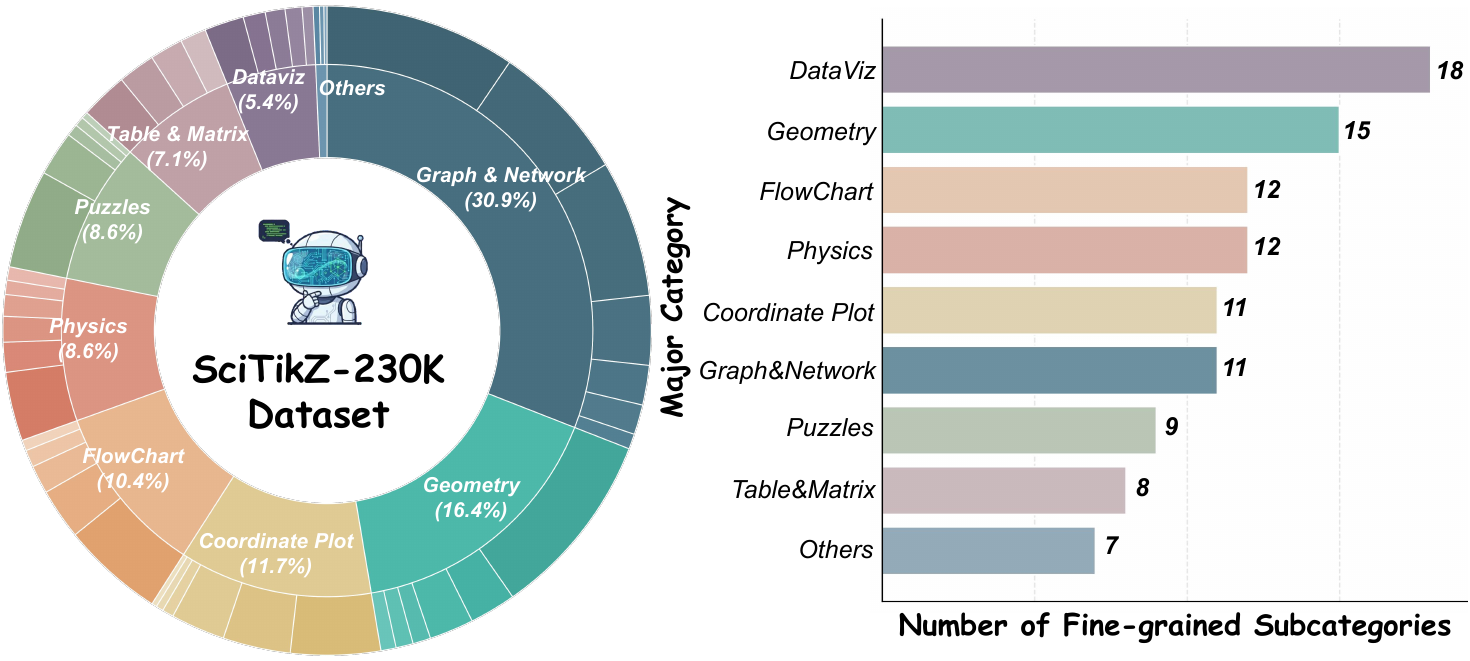}
    \caption{Overview of the SciTikZ-230K Dataset. The chart above presents the hierarchical distribution of our full dataset, spanning 11 scientific domains and over 90 fine-grained sub-categories.}
    \label{fig:dataset_stats}
\end{figure}

To bridge these gaps and cultivate intrinsic visual-coding alignment, we propose a comprehensive framework that unifies high-quality data curation with a novel RL paradigm in a closed-loop manner. \textbf{First}, we tackle the scarcity of reliable training samples by constructing a scalable Execution-Centric Data Engine. Unlike previous pipelines that passively scrape noisy web data, our engine synergizes MLLM-based semantic verification with active compilation feedback and iterative fault correction. This rigorous filtration process ensures rendering fidelity, resulting in \textbf{SciTikZ-230K} (Figure~\ref{fig:dataset_stats}), a large-scale, strictly compilable dataset that transcends the noise limitations of prior corpora to provide robust grounding for visual reasoning and executable synthesis. \textbf{Second}, we address the evaluation deficit by establishing \textbf{SciTikZ-Bench}, a multifaceted benchmark comprising 611 diverse scientific figures. This suite moves beyond simple similarity metrics, enabling the rigorous and holistic assessment of both visual fidelity and code quality across diverse scientific domains and structural patterns. \textbf{Third}, building upon this infrastructure, we introduce a novel \textbf{Dual Self-Consistency (DSC)} RL paradigm inspired by the philosophy of \textit{dual learning}\cite{he2016dual}. Addressing existing methodological constraints, DSC transcends the rigidity of SFT and the structural blindness of visual RL by establishing a robust Round-Trip Verification mechanism. Specifically, we demand that the generated code is not only visually accurate but also structurally canonical enough to be \textit{reconstructed} from its own rendered image. This \textit{self-consistency} explicitly penalizes visual hacking by suppressing the generation of degenerate, uneditable code, thereby fostering a harmony between pixel-level alignment and symbolic interpretability. Extensive experiments demonstrate that our trained models, \textbf{SciTikZer-4B/8B}, achieve state-of-the-art (SOTA) performance, significantly outperforming both general-purpose MLLMs~\cite{hurst2024gpt, comanici2025gemini, yang2025qwen3, wang2025internvl3, Qwen2.5-VL,Qwen-VL} and domain-specialized baselines~\cite{belouadi2024detikzify, saito2025sketch2diagram, zhao2025vincicoder}.

\begin{figure*}[!ht]
    \centering
    \includegraphics[width=1\linewidth]{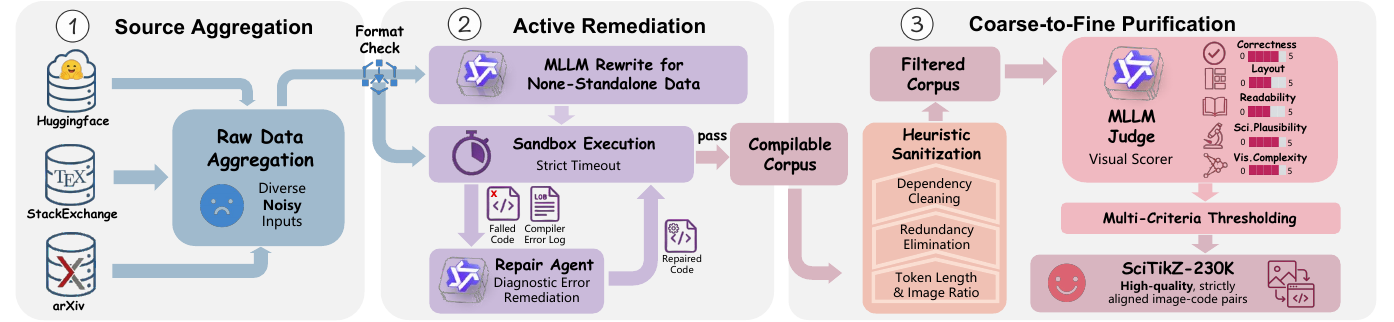}
    \caption{Overview of the \textbf{Execution-Centric Data Engine}. Beyond passive filtering, it uses MLLM-guided remediation to correct compilation faults and distill misaligned pairs, preserving diversity under strict visual-program alignment.}
    \label{fig:data_pipeline}
\end{figure*}

Specifically, our contributions are organized as follows: 
\begin{itemize} 
\item We curate \textbf{SciTikZ-230K} dataset, a high-quality, strictly compilable TikZ dataset via an Execution-Centric Data Engine, and establish \textbf{SciTikZ-Bench} for standard and comprehensive evaluation. 
\item We propose \textbf{Dual Self-Consistency (DSC)}, an RL paradigm unifying visual fidelity and structural logic through round-trip reconstruction. By forming closed-loop verification, it removes ground-truth dependency and enables logical self-consistency on unlabeled data.
\item We open-source \textbf{SciTikZer-4B/8B}. Experiments demonstrate that SciTikZer-8B achieves SOTA performance, outperforming orders-of-magnitude larger models and specialized baselines in both compilation rates and visual alignment.
\end{itemize}

\section{Related Work}
\textbf{Visual Program Synthesis.} MLLM advancements have significantly propelled visual program synthesis~\cite{khan2024self} for data visualization~\cite{seo2025vispath}. Systems like ViperGPT~\cite{suris2023vipergpt}, MatPlotAgent~\cite{yang2024matplotagent} and METAL~\cite{li2025metalmultiagentframeworkchart} drive Python libraries via an imperative paradigm of sequential commands. To standardize evaluation, benchmarks such as DePlot~\cite{liu2023deplot}, Plot2Code~\cite{wu2025plot2code}, ChartMimic~\cite{yang2024chartmimic} and ChartEdit~\cite{zhao2025chartedit} have evolved from simple extraction to complex chart reproduction and editing. Concurrently, model and data-centric advances—exemplified by ChartLlama~\cite{han2023chartllama}, ChartCoder~\cite{zhao2025chartcoder} and Chart2Code53~\cite{niu2025chart2code53}—boosted chart-to-code fidelity and scalability. However, imperative charting often abstracts geometric details. In contrast, TikZ is \textit{declarative} and compilation-critical, requiring explicit spatial specifications that challenge current MLLMs.

\textbf{Automated TikZ Generation.} Image-to-markup generation has matured in text and math domains (e.g., Im2Latex-100K~\cite{deng2017image}, Nougat~\cite{blecher2308nougat}). However, TikZ recovery fundamentally differs from image vectorization~\cite{li2020differentiable}, which yields unstructured primitives lacking the semantic topology essential for scientific diagrams. Direct TikZ synthesis remains underexplored due to data quality bottlenecks. Text-to-TikZ methods such as TikZilla~\cite{learningtikzilla} and AutomaTikZ~\cite{belouadi2023automatikz} generate compilable code from language, but struggle with image-to-TikZ, where geometry must be inferred from pixels. In this visual domain, baselines like ImgTikZ~\cite{saito2025sketch2diagram} and those using DaTikZ~\cite{belouadi2023automatikz} often rely on noisy corpora or synthetic augmentation, limiting generalization. DeTikZify~\cite{belouadi2024detikzify} enhances results via MCTS-based refinement, yet remains bounded by base model capabilities. More recently, DaVinci~\cite{xingchendavinci} uses RL with a hybrid fidelity reward to improve TikZ generation quality. In contrast, we employ DSC RL to internalize visual fidelity and structural consistency.

\textbf{Reinforcement Learning and Verifiable Generation.} RL with \emph{verifiable} feedback improves complex reasoning~\cite{lin2026mmfinereasonclosingmultimodalreasoning}, leveraging algorithms like Group Relative Policy Optimization (GRPO)~\cite{shao2024deepseekmath} specifically for code~\cite{gehring2024rlef} and math~\cite{yoshihara2025practical}. Such feedback extends to \emph{render-and-compare} supervision in RRVF~\cite{chen2025learning} and VisionR1~\cite{huang2025vision}. Related efforts leverage RL for visual reasoning: Visual Sketchpad~\cite{hu2024visual} introduces visual chain-of-thought, GRIT~\cite{fan2025grit} applies GRPO for grounded reasoning, and OpenThinkIMG~\cite{su2505openthinkimg} explores agentic policies. Closely related, RLRF~\cite{rodriguez2505rendering} optimizes SVG generation via visual rewards. Unlike RLRF's focus on SVGs, we target TikZ through DSC RL to boost both structural and visual alignment.

\begin{figure*}[t]
  \centering
  \includegraphics[width=\textwidth]{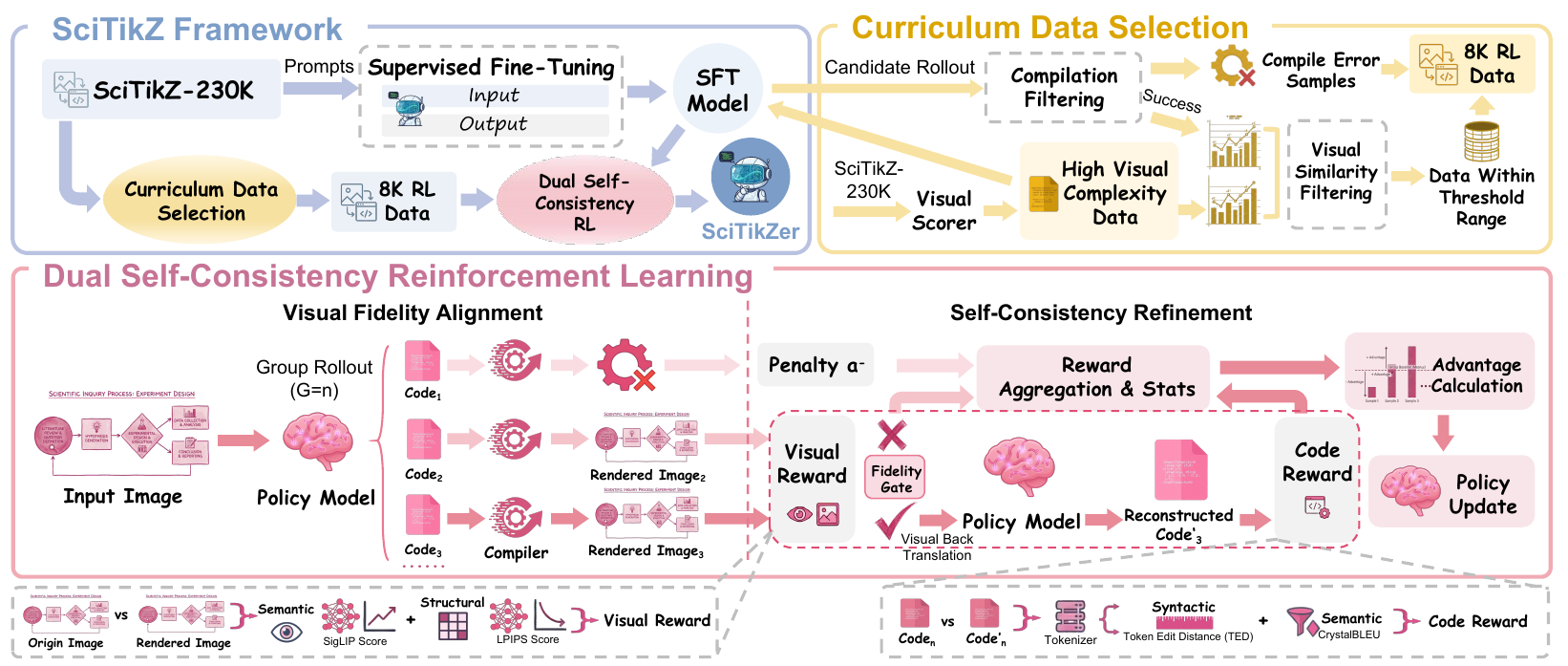}
 \caption{\textbf{Overview of the SciTikZ Framework.} The pipeline initializes with SFT and Curriculum Data Selection. Core DSC-RL integrates Visual Alignment for pixel precision and Self-Consistency Refinement via symbolic back-translation.
 }
  \label{fig:framework}
\end{figure*}

\section{SciTikZ-230K Dataset}
To curate high-quality SciTikZ-230K from heterogeneous sources, we developed an MLLM-powered data engine (Fig.~\ref{fig:data_pipeline}). This pipeline features two core components: \textbf{Active Remediation} (Sec.~\ref{subsec:active}) and \textbf{Coarse-to-Fine} Purification (Sec.~\ref{subsec:adjudication}). Final dataset statistics and additional distribution details are illustrated in Fig.~\ref{fig:dataset_stats} and Appendix.
\subsection{Source Aggregation}\label{subsec:source}
To construct a diverse data foundation, we aggregate data from HuggingFace, TeX StackExchange, and arXiv. However, direct training on these sources is impeded by three major issues, which also broadly affect most existing datasets: (i) \textbf{non-executable or non-standard code} due to missing external dependencies (e.g., \texttt{\textbackslash includegraphics}, \texttt{.bib} files), custom preamble definitions or other non-standard formatting; (ii) \textbf{visual-code misalignment}, particularly in arXiv where fragments often fail to faithfully reproduce the target; and (iii) \textbf{incomplete snippets} that introduce noise and encourage hallucinations. Instead of directly discarding suboptimal samples, our proposed Data Engine salvages and progressively refines raw corpora into more high-fidelity data, maximizing diversity through active remediation.

\subsection{Active Remediation}\label{subsec:active}
Given the low compilation success rate of raw sources and their non-standard structures (e.g., lacking \texttt{standalone} wrappers), we implement a two-stage pipeline to ensure executability and maximize data utilization at scale.

\textbf{Strict Runtime Validation.} To resolve structural inconsistencies, we employ Qwen3-VL-235B-A22B-Instruct~\cite{Qwen3-VL} to autonomously refactor the $\sim$40\% non-standalone fragments into fully self-contained and compilable formats. All blocks then undergo rigorous sandbox execution, mandating successful compilation within 10 seconds under strict constraints. This constraint ensures high-throughput and stable efficiency for the post RL phase, where rapid rendering provides a low-latency reward signal.

\textbf{Diagnostic Error Remediation.} Rather than discarding uncompilable instances, we use Qwen3-VL-235B-A22B-Instruct to iteratively repair erroneous code using compiler diagnostics. This process restores executability for approximately 60\% of faulty instances, converting them into usable training data.

\subsection{Coarse-to-Fine Purification}\label{subsec:adjudication}
After heuristic cleaning, we apply semantic validation to ensure structural uniqueness and training stability.

\textbf{Heuristic Sanitization (Coarse-grained).} We initiate our data pipeline with a cascade of lightweight heuristic filters to enforce data integrity. First, to accommodate context window limits and ensure computational efficiency during training, we discard samples with sequence lengths $\ge 8{,}192$ or image aspect ratios $> 15:1$. Next, we mitigate data redundancy using a stringent $N$-gram overlap strategy. Specifically, we calculate $50$-grams for each sample and remove those sharing more than five identical matches with the existing corpus. Finally, as a safety net, we strictly filter out any code containing unresolved external file dependencies to guarantee that every training sample is self-contained.

\textbf{Fidelity Adjudication (Fine-grained).} To ensure high-quality training targets, we adopt the robust MLLM-as-Judge paradigm. Specifically, we employ Qwen3-VL-235B-A22B-Instruct to evaluate each sample across five key dimensions (see Appendix for detail prompt): \textit{Correctness} ($s_{\text{corr}}$), \textit{Layout} ($s_{\text{lay}}$), \textit{Readability} ($s_{\text{read}}$), \textit{Scientific Plausibility} ($s_{\text{sci}}$), and \textit{Visual Complexity} ($s_{\text{comp}}$). We define the aggregate quality score as $\mathcal{S}_{\text{total}} = \sum s_i$. To curate the final dataset $\mathcal{D}_{\text{final}}$, we apply a rigorous multi-criteria filtering strategy:
\begin{equation}
\label{eq:filtering}
\begin{split}
    \mathcal{D}_{\text{final}} = \big\{ &x \in \mathcal{D}_{\text{filtered}} \mid \mathcal{S}_{\text{total}} > \delta_{\text{total}} \; \land \; s_{\text{comp}} > \delta_{\text{comp}} \\
    & \land \min(\{s_{\text{corr}}, s_{\text{lay}}, s_{\text{read}}, s_{\text{sci}}\}) > \delta_{\text{min}} \big\}.
\end{split}
\end{equation}
Following the Coarse-to-Fine Purification, we obtain \textbf{SciTikZ-230K}, a corpus of 230K high-fidelity instances characterized by superior aesthetics and precise alignment. We categorize these samples as detailed in Appendix. As visualized in Figure~\ref{fig:dataset_stats}, the dataset exhibits a diverse and well-balanced hierarchical distribution across various scientific domains. Furthermore, we establish \textbf{SciTikZ-Bench} through MLLM-based \textit{score filtering} and subsequent \textit{expert review}, ensuring visual-logical isomorphism across a stratified difficulty gradient (Easy, Medium and Hard), ranging from basic geometric primitives to more complex schematics.

\section{SciTikZer: A Faithful Img2TikZ Generator}

We introduce \textbf{SciTikZer}, a MLLM tailored for programmatic scientific graphics synthesis. As shown in Fig.~\ref{fig:framework}, our training pipeline is systematically composed of three core components: SFT on curated data, Curriculum Selection for expertise enhancement, and DSC RL for optimized visual fidelity and self-consistancy.

\subsection{Supervised Warm-up for Initialization}

We initialize the policy $\pi_\theta$ using our curated dataset $\mathcal{D}_{\text{final}}$. Formulated as a conditional sequence generation task, we optimize the model to maximize likelihood of the ground-truth code $\mathbf{y} = (y_1, \dots, y_T)$ given the input image $I$:
\begin{equation}
\label{eq:sft_loss}
  \mathcal{L}_{\text{SFT}}(\theta) = - \mathbb{E}_{(I, \mathbf{y}) \sim \mathcal{D}_{\text{final}}} \left[ \frac{1}{T} \sum_{t=1}^{T} \log \pi_\theta(y_t \mid \mathbf{y}_{<t}, I) \right].
\end{equation}
Crucially, this phase steers the model toward TikZ's strict syntax, ensuring valid compilation and providing a robust starting point for the later exploration-intensive RL phase.

\subsection{Curriculum Data Selection for RL}
To enhance sample efficiency in the RL phase, we design a curriculum that focuses on samples lying within the model's ``zone of proximal development''. We first filter $\mathcal{D}_{\text{final}}$ based on visual complexity (denoted as $s_{\text{comp}}>3$) to ensure task difficulty. We then evaluate the current policy $\pi_{\text{sft}}$ by sampling $\hat{\mathbf{y}} \sim \pi_{\text{sft}}(\cdot \mid I)$ and computing visual similarity $\mathcal{S}_{\text{vis}}$ using the SigLIP \cite{zhai2023sigmoid} encoder $\Phi_{\text{v}}$:
\begin{equation}
     \mathcal{S}_{\text{vis}}(I, \hat{\mathbf{y}}) = \frac{\Phi_{\text{v}}(I) \cdot \Phi_{\text{v}}(\mathcal{R}(\hat{\mathbf{y}}))}{\| \Phi_{\text{v}}(I) \| \| \Phi_{\text{v}}(\mathcal{R}(\hat{\mathbf{y}})) \|}.
 \end{equation}
The final RL dataset $\mathcal{D}_{\text{RL}}$ targets two categories of high-value training signals: \emph{compilation errors} and \emph{visual discrepancies}. The selection criterion is formulated as:
\begin{equation}
\label{eq:curriculum}
\scalebox{1}{ 
$
    \mathcal{D}_{\text{RL}}
    =
    \Big\{I \ \Big| \
    \text{Compile}(\hat{\mathbf{y}})=\text{Fail}
    \lor
    \big(\tau_{\min} \le \mathcal{S}_{\text{vis}}(I,\hat{\mathbf{y}}) \le \tau_{\max}\big)
    \Big\}.
$
}
\end{equation}
This strategy effectively filters out \emph{mastered} samples ($\mathcal{S}_{\text{vis}} > \tau_{\max}$) and \emph{intractable} outliers ($\mathcal{S}_{\text{vis}} < \tau_{\min}$), yielding a focused set of 8K instances that drive policy improvement. 
\subsection{Dual Self-Consistency RL}
Dual Self-Consistency RL adopts a progressive strategy to mitigate multi-objective optimization instability. We first optimize visual fidelity (\textsc{Stage 1}) to establish a renderable baseline, then incorporate self-consistency constraints in \textsc{Stage 2}. Decoupling is essential, as meaningful consistency requires visual grounding. Early enforcement without sufficient alignment risks sparse rewards and unstable convergence.
\subsubsection{Stage 1: Visual Fidelity Alignment}
While SFT captures syntax, lacking visual feedback often causes geometric misalignment. \textsc{Stage 1} employs GRPO to enforce executability and anchor policy $\pi_\theta$ via visual reward, ensuring pixel-level fidelity. This phase provides the requisite grounding for later dual-consistency optimization.

\textbf{Execution-Gated Reward.}
Rendering is defined as $\mathcal{T}(\hat{\mathbf{y}}) \rightarrow \hat{I}$, where $\hat{\mathbf{y}}$ is the generated code and $\hat{I}$ is the output image. Since invalid code yields no visual output, we construct an execution reward structure. The compilation reward $r_{\text{exec}}$ acts as a hard constraint: 
\begin{equation}
    r_{\text{exec}}(\hat{\mathbf{y}}) = 
    \begin{cases} 
        \alpha^+ & \text{if } \mathcal{T}(\hat{\mathbf{y}}) \neq \emptyset \quad (\text{Success}) \\
        \alpha^- & \text{if } \mathcal{T}(\hat{\mathbf{y}}) = \emptyset \quad (\text{Failure})
    \end{cases},
\end{equation}
where $\alpha^+ > 0$ provides a positive signal for valid syntax, and $\alpha^- \ll 0$ imposes a heavy penalty for compilation failures to prune the exploration space effectively.

\definecolor{headergray}{gray}{0.95}
\definecolor{highlightgray}{gray}{0.92}
\definecolor{posgreen}{RGB}{0, 150, 0}

\begin{table*}[!ht]
    \centering
    \fontsize{6}{7}\selectfont
    \renewcommand{\arraystretch}{0.9665}
    \setlength{\tabcolsep}{3pt}       
    
    \caption{\textbf{Main Results on SciTikZ-Bench.} Zero-shot performance comparison. Visual metrics are reported as \textbf{All / Success}, while code metrics are calculated on all samples. The best results are highlighted in \textbf{bold} and the second best are \underline{underlined}.}
    \label{tab:main_results}
    
    \resizebox{\linewidth}{!}{
        \begin{tabular}{l c c  ccccc cc}
            \toprule
            \multirow{2}{*}{\textbf{Model}} & \multirow{2}{*}{\textbf{Params}} & \textbf{Compile} & \multicolumn{5}{c}{\textbf{Visual Fidelity}} & \multicolumn{2}{c}{\textbf{Code Quality}} \\
            \cmidrule(lr){4-8} \cmidrule(lr){9-10}
             & & Success Rate$\uparrow$ & SigLIP $\uparrow$ & CLIP $\uparrow$ & LPIPS $\downarrow$ & SSIM $\uparrow$ & DreamSim $\downarrow$ & TED Norm$\downarrow$ & C-BLEU $\uparrow$ \\
            \midrule
            
            \multicolumn{10}{c}{\cellcolor{headergray}\textbf{Proprietary Models}} \\ 

            GPT-5-Mini & - & 77.3 & 72.5 / 93.7 & 70.4 / 91.1 & 54.3 / 40.9 & 55.1 / 71.4 & 33.9 / 14.5 & 53.4 & 12.7 \\
            
            Claude-4.5-Sonnet & - & 91.0 & 86.4 / 94.9 & 84.4 / 92.7 & 40.9 / 35.1 & 66.2 / 72.8 & 20.2 / 12.3 & 47.0 & 19.6 \\
            
            GPT-5.1 & - & 90.3 & 86.3 / 95.5 & 84.7 / 93.8 & 38.5 / 31.9 & 66.8 / 73.9 & 18.9 / 10.2 & 45.1 & 21.9 \\
    
            Gemini-2.5-Pro & - & 88.9 & 85.5 / 96.2 & 83.6 / 94.1 & 40.2 / 32.7 & 65.8 / 74.1 & 19.8 / 9.8 & 46.7 & 23.6 \\
            
            \multicolumn{10}{c}{\cellcolor{headergray}\textbf{Open-Source MLLMs}} \\
            DeepSeek-VL2 & 27B & 63.0 & 55.9 / 88.7 & 55.4 / 88.0 & 67.3 / 48.0 & 44.3 / 70.4 & 48.8 / 18.7 & 56.5 & 18.0 \\
            InternVL3.5-4B & 4B & 53.7 & 46.1 / 85.8 & 45.0 / 83.9 & 74.8 / 53.0 & 38.7 / 72.1 & 60.8 / 27.0 & 54.5 & 11.8 \\
            InternVL3.5-8B & 8B & 68.6 & 59.9 / 87.4 & 58.3 / 85.0 & 66.9 / 51.8 & 49.8 / 72.7 & 48.8 / 25.3 & 53.6 & 13.4 \\
            InternVL3.5-14B & 14B & 76.3 & 68.0 / 89.1 & 66.3 / 86.9 & 62.2 / 50.4 & 55.8 / 73.1 & 42.5 / 24.6 & 50.0 & 16.6 \\
            Qwen3-VL-4B & 4B & 68.1 & 61.5 / 90.3 & 60.5 / 88.9 & 64.2 / 47.4 & 49.1 / 72.0 & 45.1 / 19.4 & 51.0 & 15.9 \\
            Qwen3-VL-8B & 8B & 71.0 & 64.5 / 90.8 & 63.3 / 89.1 & 60.7 / 44.6 & 50.9 / 71.7 & 42.3 / 18.8 & 50.8 & 17.0 \\
            Qwen3-VL-32B & 32B & 82.8 & 77.6 / 93.7 & 75.6 / 91.3 & 52.6 / 42.7 & 59.3 / 71.6 & 28.2 / 13.3 & 50.1 & 19.1 \\
            Qwen3-VL-235B-A22B & 235B & 92.1 & 86.8 / 94.2 & 84.8 / 92.0 & 43.8 / 39.0 & 66.6 / 72.2 & 19.2 / 12.3 & 45.2 & 23.3 \\
            
            \multicolumn{10}{c}{\cellcolor{headergray}\textbf{Task-Specific MLLMs}} \\
             ImgTikZ & 8B & 81.8 & 76.8 / 93.9 & 75.4 / 92.2 & 49.9 / 38.8 & 58.9 / 72.0 & 27.3 / 11.2 & 48.3 & 20.2 \\
             VinciCoder-8B & 8B & 83.6 & 78.2 / 93.4 & 76.3 / 91.2 & 50.0 / 40.2 & 60.4 / 72.3 & 26.3 / 11.9 & 47.8 & 23.8 \\
             DeTikZify-CL-7B & 7B & 77.1 & 70.1 / 91.0 & 68.8 / 89.3 & 56.5 / 43.6 & 54.9 / 71.2 & 36.4 / 17.5 & 47.6 & 19.5 \\
             DeTikZify-DS-7B & 7B & 77.4 & 70.8 / 91.4 & 69.5 / 89.8 & 56.0 / 43.2 & 55.3 / 71.4 & 36.6 / 18.1 & 49.3 & 18.9 \\
            DeTikZify-V2-8B & 8B & 85.3 & 80.3 / 94.1 & 78.9 / 92.5 & 42.2 / 32.2 & 65.3 / \textbf{76.6} & 23.2 / 9.9 & \underline{42.9} & 28.8 \\
            DeTikZify-V2.5-8B & 8B & 93.1 & 88.9 / 95.4 & 87.1 / 93.6 &  37.3 / 32.7 & 70.4 / \underline{75.6} & 15.9 / 9.7 & 43.0 & \textbf{30.4} \\
            
            \rowcolor{highlightgray} \textbf{SciTikZer-4B (Ours)} & \textbf{4B} & \underline{95.9} & \underline{92.4} / \underline{96.3} & \underline{90.6} / \underline{94.4} & \underline{30.8} / \underline{27.9} & \underline{70.7} / 73.8 & \underline{12.2} / \underline{8.5} & 43.2 & 28.6 \\
            
             \rowcolor{highlightgray} $\Delta$ vs Qwen3-VL-4B & - & 
            \gain{\uparrow}{27.8} & 
            \gain{\uparrow}{30.9} / \gain{\uparrow}{6.0} & 
            \gain{\uparrow}{30.1} / \gain{\uparrow}{5.5} & 
            \gain{\downarrow}{33.4} / \gain{\downarrow}{19.5} & 
            \gain{\uparrow}{21.6} / \gain{\uparrow}{1.8} & 
            \gain{\downarrow}{32.9} / \gain{\downarrow}{10.9} & 
            \gain{\downarrow}{7.8} & 
            \gain{\uparrow}{12.7} \\
             
            \rowcolor{highlightgray} \textbf{SciTikZer-8B (Ours)} & \textbf{8B} & \textbf{97.2} & \textbf{93.8} / \textbf{96.5} & \textbf{92.3 / 94.9} & \textbf{29.7 / 27.7} & \textbf{72.5} / 74.6 & \textbf{10.9} / \textbf{8.4} & \textbf{42.8} & \underline{28.9} \\

            \rowcolor{highlightgray} $\Delta$ vs Qwen3-VL-8B & - & 
            \gain{\uparrow}{26.2} & 
            \gain{\uparrow}{29.3} / \gain{\uparrow}{5.7} & 
            \gain{\uparrow}{29.0} / \gain{\uparrow}{5.8} & 
            \gain{\downarrow}{31.0} / \gain{\downarrow}{16.9} & 
            \gain{\uparrow}{21.6} / \gain{\uparrow}{2.9} & 
            \gain{\downarrow}{31.4} / \gain{\downarrow}{10.4} & 
            \gain{\downarrow}{8.0} & 
            \gain{\uparrow}{11.9} \\
           
            \bottomrule
        \end{tabular}
    }
\end{table*}

\textbf{Multi-Granularity Visual Alignment Reward.}
Upon successful compilation, we compute $r_{\text{vis}}$ via a dual-stream approach. Departing from single-latent similarity in prior works, our method fuses high-level semantics with low-level perceptual structure to ensure multi-granular alignment. 

\textit{1) Semantic Alignment (SigLIP):} We employ the SigLIP encoder $\Phi_{\text{sem}}$ to ensure a semantic match with the source. To mitigate reward hacking on trivial backgrounds, we introduce {hinge-scaled similarity}. Defining $s_{\text{raw}} = \cos(\Phi_{\text{sem}}(I), \Phi_{\text{sem}}(\hat{I}))$, the visual score of semantic alignment is:
\begin{equation} 
 s_{\text{sem}}(I, \hat{I}) = \frac{\max\left(0, s_{\text{raw}} - \tau_{\text{hold}}\right)}{1 - \tau_{\text{hold}}}, 
\end{equation}
where $\tau_{\text{hold}}$ is the baseline fidelity threshold. This renormalization suppresses low-quality noise while amplifying gradients for high-fidelity samples, sharpening the distinction between roughly correct and precisely aligned outputs.

\textit{2) Structural Precision (LPIPS \cite{zhang2018unreasonable}):} Structural accuracy is critical during early optimization to prevent layout collapse, often missed by semantic encoders. We employ LPIPS with an AlexNet \cite{krizhevsky2012imagenet} backbone ($\Phi_{\text{alex}}$) to enforce fine-grained geometric precision. Distance $d_{\text{lpips}}$ is mapped to a normalized score via an exponential kernel:
\begin{equation}
    s_{\text{struct}}(I, \hat{I}) = \exp\left( - \frac{d_{\text{lpips}}(I, \hat{I})}{\tau_{\text{temp}}} \right),
\end{equation}
where $\tau_{\text{temp}}$ controls the sensitivity of the spatial penalty.

\textbf{Optimization via GRPO.}
We synthesize semantic and structural feedback into a unified visual reward $r_{\text{vis}} = \lambda_1 s_{\text{sem}} + \lambda_2 s_{\text{struct}}$. The total reward $R_{\text{vis}}(I, \hat{\mathbf{y}})$ integrates the hard execution constraint via a compilation gate:
\begin{equation}
    R_{\text{vis}}(I, \hat{\mathbf{y}}) = r_{\text{exec}}(\hat{\mathbf{y}}) + \mathbf{1}_{\{\mathcal{T}(\hat{\mathbf{y}}) \neq \emptyset\}} \cdot \lambda_{\text{vis}} \cdot r_{\text{vis}}(I, \mathcal{T}(\hat{\mathbf{y}})),
\end{equation}
where $\mathbf{1}$ is the indicator function for successful compilation, and $\lambda_{\text{vis}}$ balances the visual feedback scale.
To optimize the policy $\pi_\theta$ efficiently, we employ GRPO. For each input $I$, we sample a group of $G$ outputs $\{\hat{\mathbf{y}_1}, \dots, \hat{\mathbf{y}_G}\}$ from the old policy $\pi_{\theta_{\text{old}}}$. The optimization objective is:
\begin{equation}
\begin{aligned}
    \mathcal{J}_{\text{GRPO}}(\theta) &= \mathbb{E}_{I \sim \mathcal{D}} \bigg[ \frac{1}{G} \sum_{i=1}^G \min \Big( \rho_i \hat{A}_i, \text{clip}(\rho_i, 1{-}\epsilon, 1{+}\epsilon) \hat{A}_i \Big) \\
    &\quad - \beta \mathbb{D}_{\text{KL}}(\pi_\theta \| \pi_{\text{ref}}) \bigg],
\end{aligned}
\end{equation}
where $\rho_i = \frac{\pi_\theta(\hat{\mathbf{y}_i}|I)}{\pi_{\theta_{\text{old}}}(\hat{\mathbf{y}_i}|I)}$ denotes the probability ratio, and $\beta$ scales the KL-penalty against $\pi_{\text{ref}}$. GRPO stabilizes training by estimating the advantage $\hat{A}_i$ via in-group normalization:
\begin{equation}
    \hat{A}_i = \frac{R(I,\hat{\mathbf{y}_i}) - \mu_{\{R\}}}{\sigma_{\{R\}}},
\end{equation}
where $\mu_{\{R\}}, \sigma_{\{R\}}$ denote the group reward statistics.
\subsubsection{Stage 2: Self-Consistency Refinement}
While \textsc{Stage 1} establishes visual grounding, structural constraints alone cannot guarantee canonical code generation. To boost logical robustness, \textsc{Stage 2} introduces a symbolic round-trip mechanism inspired by dual learning~\cite{he2016dual}. Unlike multi-agent systems, we leverage the deterministic compiler $\mathcal{T}$ to form a closed-loop feedback system within a single policy, internalizing structural reciprocity.

\textbf{Dual Consistency Formulation.}
Given an image $I$, the policy generates a code sequence $\hat{\mathbf{y}} \sim \pi_\theta(\cdot|I)$. This code is rendered into a synthetic image $\hat{I} = \mathcal{T}(\hat{\mathbf{y}})$. Subsequently, we query the same policy to back-translate the synthetic image into a reconstructed code $\hat{\mathbf{y}}' \sim \pi_\theta(\cdot|\hat{I})$. This formulation couples dual directions, so that round-trip consistency under compiler-verified rendering provides additional supervision and structural constraints.
The intuition is that if the model truly understands the visual syntax, the code generated from its own rendering ($\hat{\mathbf{y}}'$) should be \textbf{structurally consistent} with the original code ($\hat{\mathbf{y}}$), i.e., $\hat{\mathbf{y}} \approx \hat{\mathbf{y}}'$.

\textbf{Composite Structural-Semantic Reward.}
Formatting noise hinders string-based quantification of deviations between the primal $\hat{\mathbf{y}}$ and reconstructed $\hat{\mathbf{y}}'$. We thus define a score $s_{\text{code}}$ unifying structural topology and semantics.

\textit{1) Kernelized Token Edit Distance (TED):}
We tokenize the \LaTeX{} source using a domain-specific lexer, $\hat{\mathbf{t}}=\mathcal{L}(\hat{\mathbf{y}})$, and compute the Extended Edit Distance (EED)~\cite{stanchev2019eed} between token streams. We then normalize the unbounded distance $\mathcal{D}_{\text{eed}}$ using a Gaussian kernel:

\begin{equation}
    s_{\text{ted}}(\hat{\mathbf{y}}, \hat{\mathbf{y}}') = \exp\left( - \frac{\mathcal{D}_{\text{eed}}(\mathcal{L}(\hat{\mathbf{y}}), \mathcal{L}(\hat{\mathbf{y}}'))}{\tau_{\text{ted}}} \right),
\end{equation}
where temperature $\tau_{\text{ted}}$ regulates structural sensitivity.

\textit{2) CrystalBLEU with Frequency Masking:}
To mitigate boilerplate inflation from \LaTeX{} syntax, we employ CrystalBLEU \cite{eghbali2022crystalbleu} to isolate semantic fidelity. We suppress a pre-computed set $\mathcal{T}_k$ of frequent n-grams from the corpus. The refined precision $p_n^*$ uses an indicator function to filter redundant trivial syntax:
\begin{equation}
    p_n^* = \frac{\sum_{g \in \mathcal{G}_n(\mathbf{y}')} \mathbf{1}_{\{g \notin \mathcal{T}_k\}} \cdot \min\left(C(g|\hat{\mathbf{y}}'), C(g|\hat{\mathbf{y}})\right)}{\sum_{g \in \mathcal{G}_n(\hat{\mathbf{y}}')} \mathbf{1}_{\{g \notin \mathcal{T}_k\}} \cdot C(g|\hat{\mathbf{y}}')},
\end{equation}
where $\mathbf{1}_{{g \notin \mathcal{T}_k}}$ excludes frequent templates. Their convex combination defines the code-consistency reward:
\begin{equation}
\scalebox{1}{$
    s_{\text{code}}(\hat{\mathbf{y}}, \hat{\mathbf{y}}') = \gamma \cdot \text{CrystalBLEU}(\hat{\mathbf{y}}, \hat{\mathbf{y}}') + (1-\gamma) \cdot s_{\text{ted}}(\hat{\mathbf{y}}, \hat{\mathbf{y}}')$},
\end{equation}
where $\gamma$ prioritizes semantic fidelity over raw syntax.

\textbf{Fidelity-Gated Optimization.}
A critical risk in self-supervised training is \textit{mode collapse}, where degenerate code $\hat{\mathbf{y}}$ yields trivial $\hat{I}$ that easily maps back. To prevent reinforcing such loops, we introduce a \textbf{Fidelity-Gated} mechanism. The self-consistency reward activates only when the intermediate visual alignment exceeds a threshold $\tau_{\text{gate}}$:
\begin{equation}
\scalebox{1}{$
    R_{\text{total}}(\mathbf{y}, I) = R_{\text{vis}}(I, \hat{\mathbf{y}}) + \mathbf{1}_{[r_{\text{vis}} > \tau_{\text{gate}}]} \cdot \lambda_{\text{code}} \cdot s_{\text{code}}(\hat{\mathbf{y}}, \hat{\mathbf{y}}')$}.
\end{equation}
This acts as a quality filter, rewarding self-consistency \textit{only if} the primary generation is sufficiently visually faithful. A detailed illustration of the workflow of DSC-RL is provided in the Appendix.

\section{Experiments}
In this section we evaluates SciTikZer and answer our key research questions (RQs). We begin by detailing the experimental setup in Sec.~\ref{sec:experiments_setup}, followed by analyses of the main results in Sec.~\ref{sec:main_results}. We then conduct training analysis and ablation studies in Sec.~\ref{sec:progressive_training} and Sec.~\ref{sec:ablation_study}. Finally, we assess the cross-language generalization of DSC to Python (Sec.~\ref{sec:generalization}) and present human/case analyses (Sec.~\ref{sec:case_analysis}). Additional details are provided in Appendix.

\definecolor{headergray}{gray}{0.95}
\definecolor{highlightgray}{gray}{0.92}

\begin{table*}[!ht]
    \centering
    \caption{\textbf{Progressive Training Framework Analysis on SciTikZ-Bench.} This table disentangles the cumulative gains from each training stage: Base model, SFT, Stage 1 RL (Visual Alignment), and the full SciTikZer with Stage 2 DSC-RL.}
    \label{tab:progressive_analysis}
    \resizebox{\linewidth}{!}{
        \begin{tabular}{l l c c c c c c c c}
            \toprule
            \textbf{Scale} & \textbf{Training Stage} & \textbf{Compile Rate}$\uparrow$ & \textbf{SigLIP}$\uparrow$ & \textbf{CLIP}$\uparrow$ & \textbf{LPIPS}$\downarrow$ & \textbf{SSIM}$\uparrow$ & \textbf{DreamSim}$\downarrow$ & \textbf{TED Norm}$\downarrow$ & \textbf{C-BLEU}$\uparrow$ \\
            \midrule
            
            \multirow{4}{*}{4B} & Base (Qwen3-VL-4B) & 68.1 & 61.5 / 90.3 & 60.5 / 88.9 & 64.2 / 47.4 & 49.1 / 72.0 & 45.1 / 19.4 & 51.0 & 15.9 \\
                                & + SFT & 80.4 & 74.3 / 92.5 & 72.4 / 90.1 & 52.5 / 40.9 &  58.8 / 73.2 & 32.8 / 16.4 & 45.6 & \textbf{30.7} \\
                                & + SFT + Stage 1 & \underline{90.7} & \underline{86.0} / \underline{94.8} & \underline{84.0} / \underline{92.6} & \underline{37.5} / \underline{31.1} & \underline{66.7} / \underline{73.6} &  \underline{19.4} / \underline{11.1} & \underline{44.1} & \underline{29.1} \\
            \rowcolor{highlightgray} & \textbf{+ SFT + Stage 1 + Stage 2} & \textbf{95.9} & \textbf{92.4} / \textbf{96.3} & \textbf{90.6} / \textbf{94.4} & \textbf{30.8} / \textbf{27.9} & \textbf{70.7} / \textbf{73.8} & \textbf{12.2} / \textbf{8.5} & \textbf{43.2} & 28.6 \\
            
            \midrule
            
            \multirow{4}{*}{8B} & Base (Qwen3-VL-8B) & 71.0 & 64.5 / 90.8 & 63.3 / 89.1 & 60.7 / 44.6 & 50.9 / 71.7 & 42.3 / 18.8 & 50.8 & 17.0 \\
                                & + SFT & 81.0 & 75.1 / 92.7 & 73.3 / 90.5 & 51.4 / 40.0 & 59.8 / 73.8 & 30.3 / 14.0 & 45.2 & \textbf{31.6} \\
                                & + SFT + Stage 1 & \underline{91.2} & \underline{86.3} / \underline{94.7} & \underline{85.2} / \underline{93.5} & \underline{37.0} / \underline{30.9} & \underline{67.6} / \underline{74.1} &  \underline{18.8} / \underline{10.9} & \textbf{42.6} & 28.7 \\
            \rowcolor{highlightgray} & \textbf{+ SFT + Stage 1 + Stage 2} & \textbf{97.2} & \textbf{93.8} / \textbf{96.5} & \textbf{92.3} / \textbf{94.9} & \textbf{29.7} / \textbf{27.7} & \textbf{72.5} / \textbf{74.6} & \textbf{10.9} / \textbf{8.4} & \underline{42.8} & \underline{28.9} \\
            
            \bottomrule
        \end{tabular}
    }
\end{table*}

\subsection{Experimental Setup}
\label{sec:experiments_setup}
\textbf{Training details.}
We use LLaMA-Factory~\cite{zheng2024llamafactory} for supervised fine-tuning and EasyR1~\cite{zheng2025easyr1}, built upon \texttt{verl}~\cite{sheng2024hybridflow}, for reinforcement learning. 
For SFT, the learning rate is set to $5\times10^{-5}$, with a total batch size of 128, a maximum token length of 4096, and 3 training epochs. 
For RL, we adopt AdamW~\cite{Loshchilov2019AdamW} and conduct all experiments on $8\times$ NVIDIA A100 (80GB) GPUs. 

\textbf{Baselines.} We compare SciTikZer against three groups: proprietary MLLMs, including GPT-5-Mini and GPT-5.1~\cite{hurst2024gpt}, Gemini-2.5-Pro~\cite{comanici2025gemini}, and Claude-4.5-Sonnet~\cite{anthropic_sonnet45_system_card}; open-weight MLLMs, including InternVL3.5~\cite{wang2025internvl3}, DeepSeek-VL2~\cite{liu2024deepseekvl2}, and Qwen3-VL-Instruct~\cite{yang2025qwen3}; and task-specific models, including DeTikZify~\cite{belouadi2024detikzify}, ImgTikZ~\cite{saito2025sketch2diagram}, and VinciCoder~\cite{zhao2025vincicoder}. Together, these baselines provide a comprehensive comparison across model scales.

\textbf{Benchmarks.} To address the lack of comprehensive benchmarks in this domain, we introduce \textbf{SciTikZ-Bench}, comprising 611 manually verified and decontaminated samples. This benchmark enables a multi-dimensional assessment, ranging from visual fidelity to code quality. To ensure robust evaluation, we additionally benchmark our model on the established DaTikZ-v3~\cite{belouadi2023automatikz} test set. 

\textbf{Evaluation Metrics.} Our protocol covers two aspects: (1) \textbf{Visual Fidelity}: We measure semantic alignment via SigLIP/CLIP, and structural precision via LPIPS, SSIM, and DreamSim\cite{fu2023dreamsim}. On DaTikZ-v3, we also include KID ($\times 10^3$) following standard practice. (2) \textbf{Code Quality}: We use CrystalBLEU (cBLEU) to assess token overlap excluding boilerplate. To measure structural divergence, we use Token Edit Distance (TED)~\cite{stanchev2019eed} for SciTikZ-Bench. For DaTikZ-v3, we instead follow the evaluation protocol used in prior work and report the baseline-specific TeX Edit Distance~\cite{belouadi2024detikzify}, ensuring fair comparison with previously reported results.

\subsection{Main Results}
\label{sec:main_results}
\textbf{RQ1: How well does SciTikZer models perform on SciTikZ-Bench?} Table~\ref{tab:main_results} presents the comprehensive evaluation results, where SciTikZer-8B establishes a new SOTA with a near-perfect 97.2\% compilation success rate. It significantly outperforms both proprietary giants like Gemini-2.5-Pro (88.9\%) and massive open-source models like Qwen3-VL-235B (92.1\%), demonstrating the high efficiency of our data-centric fine-tuning. Furthermore, compared to specialized baselines, SciTikZer-8B extends its lead. While the recent top-performing DeTikZify-V2.5-8B achieves a competitive 93.1\%, our model shows a decisive advantage in visual fidelity metrics. As shown in Table~\ref{tab:main_results}, SciTikZer-8B surpasses DeTikZify-V2.5-8B by a clear margin in semantic alignment (SigLIP: 93.8 vs. 88.9) and structural precision (LPIPS: 29.7 vs. 37.3). Although DeTikZify-V2.5 shows a marginal edge in C-BLEU (30.4 vs. 28.9), our model achieves a lower Token Edit Distance (TED: 42.8 vs. 43.0). This indicates that SciTikZer prioritizes generating code that renders visually accurate figures rather than merely maximizing n-gram overlap, making it a robust tool for real-world scientific illustration.

\textbf{RQ2: How well does SciTikZer models generalize to other datasets?} To verify that our model has learned generalized syntax rather than overfitting, we report results on the external DaTikZ-v3 test set in Table~\ref{tab:datikz_eval_single}. SciTikZer-8B maintains its dominance across all metrics, achieving the highest compilation rate (94.46\%) and the lowest distribution gap (KID: 1.14). Notably, it outperforms the 235B Qwen3-VL model in both code quality (cBLEU: 16.17 vs. 16.05) and visual similarity (DSim: 88.29 vs. 83.91). These results confirm that SciTikZer possesses strong generalization capabilities, effectively handling diverse TikZ styles beyond its training distribution. This robustness underscores the effectiveness of our data construction and reinforcement learning framework in supporting reliable out-of-distribution graphics program synthesis.

\begin{table}[t]
    \centering
    \caption{\textbf{Performance on DaTikZ-v3 test set.} The best results are highlighted in \textbf{bold} and the second best are \underline{underlined}.}
    \renewcommand{\arraystretch}{1}
    \setlength{\tabcolsep}{3pt}
    \footnotesize 
    \resizebox{\linewidth}{!}{%
        \begin{tabular}{l c c c c c}
            \toprule
            \textbf{Model} & \textbf{DSim} $\uparrow$ & \textbf{KID} $\downarrow$ & \textbf{cBLEU} $\uparrow$ & \textbf{TED} $\downarrow$ & \textbf{Compile} $\uparrow$ \\
            \midrule
            
            Gemini-2.5-Pro & \underline{87.34} & 2.86 & 12.29 & 50.45 & 66.42 \\
            
            Qwen3-VL-8B & 78.25 & 5.32 & 7.75 & 60.39 & 55.17 \\
            InternVL3.5-14B & 72.51 & 7.70 & 10.53 & 53.45 & 65.31 \\
            Qwen3-VL-235B-A22B & 83.91 & 2.66 & \underline{16.05} & \underline{49.80} & 87.82 \\
            
            DeTikZify-v2.5-8b & 85.05 & \underline{1.72} & 12.83 & 51.10 & \underline{90.41} \\
            
            \midrule
            
            \rowcolor{highlightgray}
            \textbf{SciTikZer-8B (Ours)} & \textbf{88.29} & \textbf{1.14} & \textbf{16.17} & \textbf{48.83} & \textbf{94.46} \\
            
            \bottomrule
        \end{tabular}%
    }
    
    \label{tab:datikz_eval_single}
\end{table}

\subsection{Analysis of Progressive Training}
\label{sec:progressive_training}
\textbf{RQ3: What is the contribution of each stage in the progressive training pipeline?}
We evaluate the cumulative impact of our training pipeline across four stages for both 4B and 8B scales. The initial SFT phase establishes a solid syntactic foundation, nearly doubling C-BLEU and improving compilation rates by approximately 10\% over the base models. The subsequent Stage 1 (Visual RL) yields the most substantial leap in visual fidelity—notably slashing the 8B model's LPIPS from 51.4 to 37.0—demonstrating that direct render-based feedback is essential for precise geometric grounding. Finally, Stage 2 (DSC-RL) provides critical structural refinement, pushing the 8B compilation rate to a peak of 97.2\% and further optimizing fine-grained perceptual metrics like SSIM and DreamSim. Crucially, the slight trade-off in C-BLEU during Stage 2, paired with improved visual alignment, suggests that our DSC mechanism successfully steers the model away from lexical overfitting and visual hacking toward more robust, logically self-consistent program synthesis.

\subsection{Ablation Study}
\label{sec:ablation_study}

\begin{figure}[t]
    \centering
    \includegraphics[width=\linewidth]{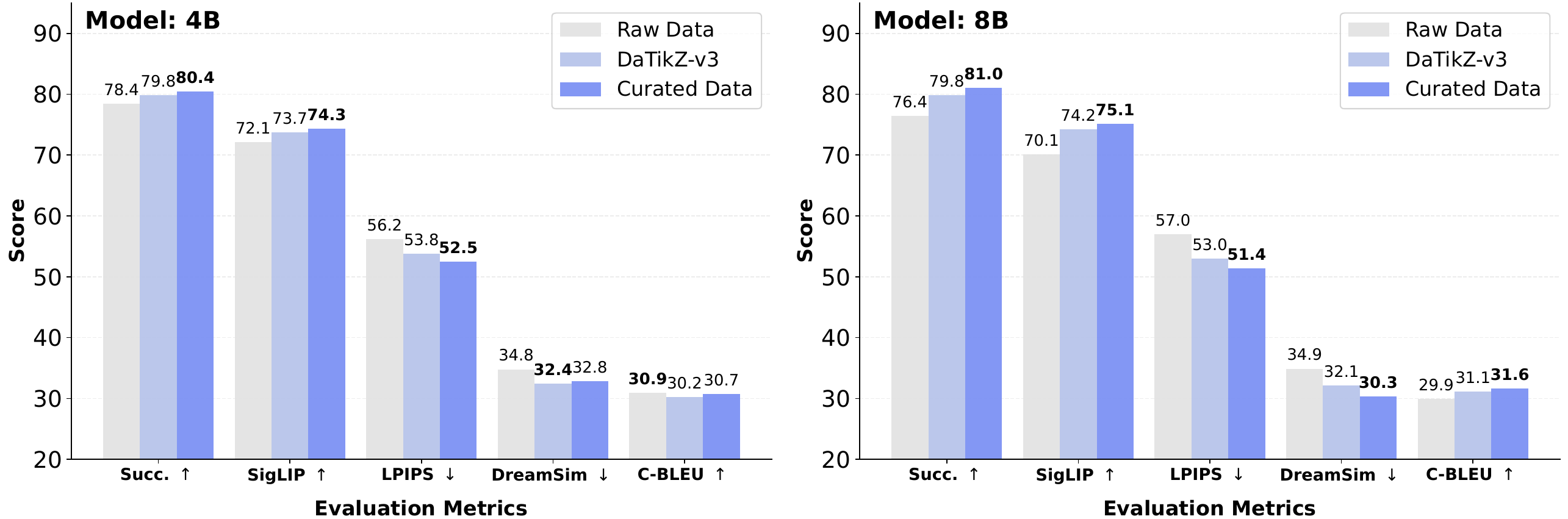}
    \caption{\textbf{Impact of Data Curation.} Curated data enhances both executability and visual fidelity in 4B and 8B models compared with raw data and DaTikZ-v3.}
    \label{fig:ablation_data}
\end{figure}

\begin{figure}[t]
    \centering
\includegraphics[width=\linewidth]{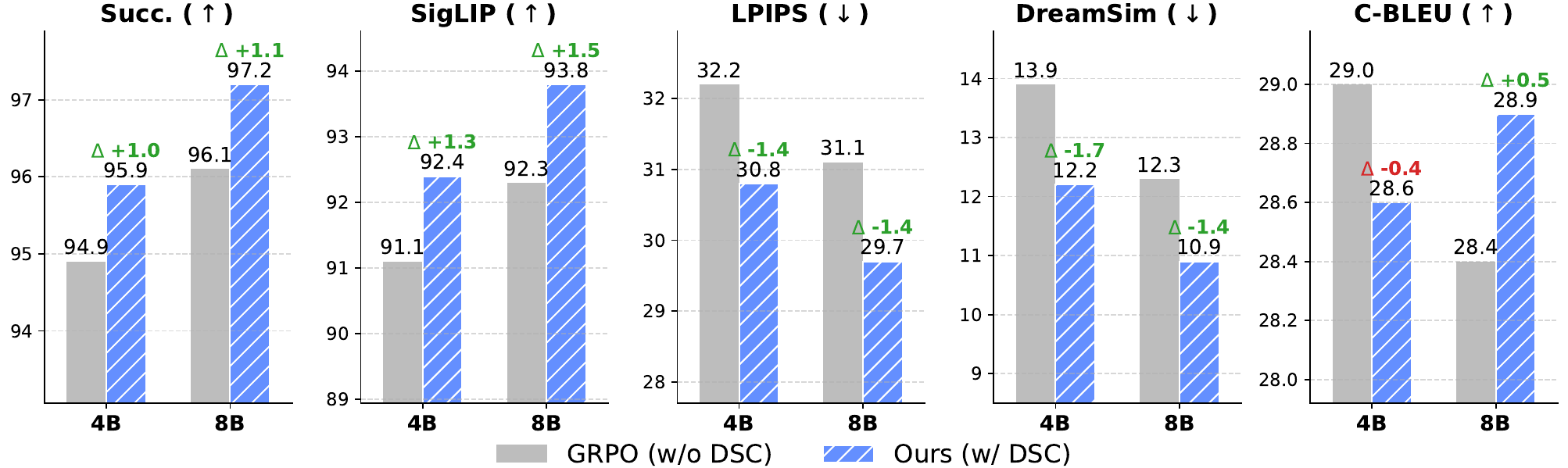}
   \caption{\textbf{Ablation of Dual Self-Consistency (DSC).} Comparison between standard RL and our full DSC method.}
    \label{fig:ablation_dsc}
\end{figure}

\begin{figure}[t]
  \centering
  \includegraphics[width=\linewidth]{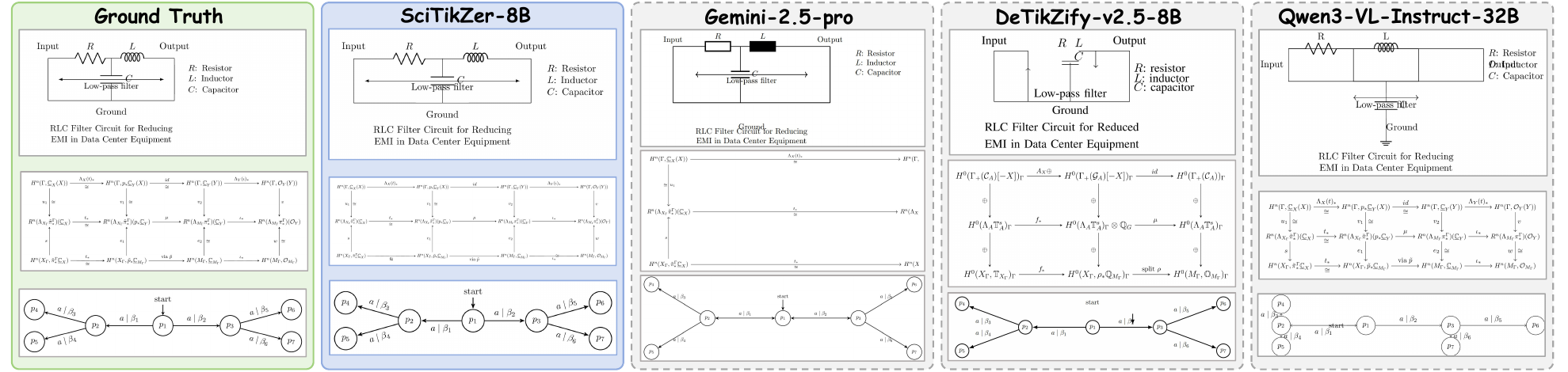}
  \caption{\textbf{Case Analysis.} A qualitative comparison of rendered
  TikZ code generated by different models.
  We compare SciTikZer-8B with SOTA baselines, including
  Gemini-2.5-Pro, DeTikZify-v2.5-8B, and Qwen3-VL-Instruct-32B.}
  \label{fig:case_analysis}
\end{figure}

\textbf{RQ4: How much does data curation matter?} As illustrated in Figure~\ref{fig:ablation_data}, training on the curated SciTikZ-230K yields consistent improvements across model scales. For the 8B model, curation boosts the compilation rate from 76.4\% to 81.0\% and increases the SigLIP score by +5.0 points. Compared with DaTikZ-v3, SciTikZ-230K performs better on most metrics, further validating our execution-centric curation strategy. This substantial gain confirms that ensuring data quality enable model to learn the intricate mapping between visual layouts and TikZ code more effectively.

\textbf{RQ5: Does Dual Self-Consistency (DSC) improve performance?} Figure \ref{fig:ablation_dsc} compares the standard GRPO baseline and our full DSC framework. Across both model scales, the integration of DSC yields measurable improvements in fidelity. For the 8B model, DSC elevates the compilation success rate to 97.2\% and reduces the LPIPS perceptual distance from 30.8 to 29.7. Notably, while GRPO alone sometimes leads to higher C-BLEU (e.g., 29.0 vs. 28.6 in the 4B scale), DSC prioritizes structural consistency, resulting in better visual alignment. This indicates that the DSC constraint helps the model maintain logical self-consistency rather than solely optimizing for superficial rendering success.

\subsection{Cross-Language Generalization} \label{sec:generalization} \textbf{RQ6: Can DSC RL generalize across programming languages?} To verify the versatility of our approach, we applied the Dual Self-Consistent RL framework to Python generation, moving beyond declarative TikZ. We utilized \textbf{VinciCoder-8B-SFT}~\cite{zhao2025vincicoder} as the backbone and evaluated on ChartMimic~\cite{yang2024chartmimic}. As shown in Table~\ref{tab:chartmimic}, incorporating DSC into the self-consistency structural reward mechanism effectively regularizes the imperative action space. Consequently, our method outperforms standard RL baselines in both executability and visual fidelity, confirming its robustness across different coding languages.
\begin{table}[t]
    \centering
    \setlength{\tabcolsep}{10pt}
    \renewcommand{\arraystretch}{0.6}
    \caption{\textbf{Performance on ChartMimic\_direct\_v2.} Comparison against standard VinciCoder RL method.}
    \label{tab:chartmimic}
    
    \resizebox{\linewidth}{!}{
        \begin{tabular}{l c c c}
            \toprule
            \multirow{2}{*}{\textbf{Model}} & \multicolumn{3}{c}{\textbf{ChartMimic\_direct\_v2}} \\
            \cmidrule(lr){2-4}
             & \textbf{Exec. Rate}$\uparrow$ & \textbf{Low-L}$\uparrow$ & \textbf{High-L}$\uparrow$ \\
            \midrule
            \fontsize{9}{10}\selectfont
            VinciCoder-8B-SFT &  \fontsize{9}{10}\selectfont 87.9 &  \fontsize{9}{10}\selectfont 75.2 &  \fontsize{9}{10}\selectfont 79.3 \\
             \fontsize{9}{10}\selectfont VinciCoder-8B-RL &  \fontsize{9}{10}\selectfont 91.2 &  \fontsize{9}{10}\selectfont 79.3 &  \fontsize{9}{10}\selectfont 80.9 \\
            \rowcolor{highlightgray}
             \fontsize{9}{10}\selectfont \textbf{VinciCoder-8B-DSC} &  \fontsize{9}{10}\selectfont \textbf{92.1} &  \fontsize{9}{10}\selectfont \textbf{80.2} & \textbf{81.5} \\
            \bottomrule
        \end{tabular}
    }
\end{table}

\begin{table}[t]
    \centering
    \includegraphics[width=\linewidth]{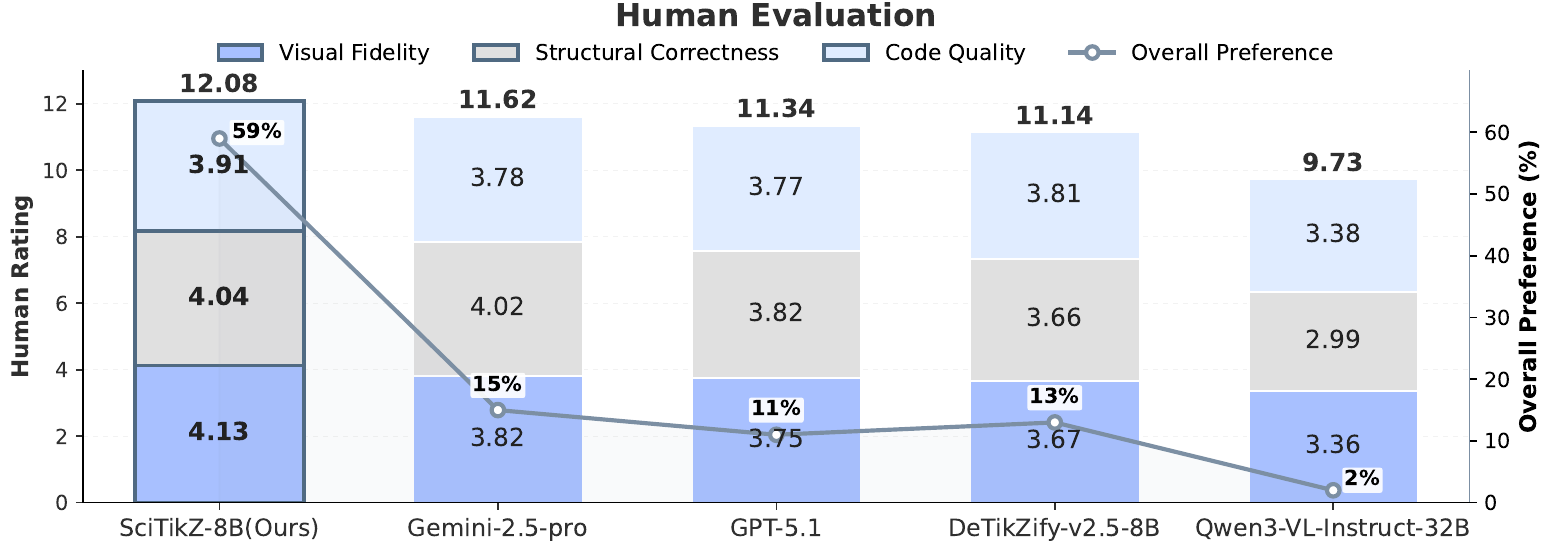}
    \caption{Blind human evaluation carefully conducted on 300 randomly sampled benchmark test examples.}
    \label{tab:human_eval}
\end{table}

\subsection{Human Evaluation and Case Analysis}
\textbf{RQ7: Does our method improve human-perceived quality and structural fidelity?}
\label{sec:case_analysis}
To complement automatic metrics, we conduct a blind human evaluation with six annotators on 300 randomly sampled examples. We retain only examples for which all four systems produce compilable outputs and present each reference image with four anonymized candidates in random order. Annotators select the best candidate and rate each output on a 1--5 Likert scale for \textit{Visual Fidelity}, \textit{Structural Correctness}, and \textit{Code Quality}. Detailed instructions are provided in the appendix. 

As shown in Table~\ref{tab:human_eval}, our trained model SciTikZer-8B is the most preferred model in human evaluation, receiving the highest overall preference (59\%) and the best aggregate human score (12.08), with a clear margin over the strongest baseline. This indicates that dual self-consistency training improves not only overall visual quality but also structural fidelity and code quality. The consistent gains across all dimensions further demonstrate the robustness and practical utility of our approach for scientific graphics synthesis.

We further conduct case analysis across multiple representative models. As shown in Figure~\ref{fig:case_analysis}, representative examples from the benchmark test set consistently show that, compared with strong baselines, SciTikZer-8B more accurately captures complex structural details, coordinate alignment, and fine-grained spatial relations, leading to better visual consistency and logical coherence.

\section{Conclusion}
In this paper, we address graphics program synthesis by enabling MLLMs to generate TikZ code for scientific figures. We introduce SciTikZ-230K, a large-scale scientific graphics dataset, and SciTikZ-Bench, a multifaceted evaluation benchmark. By proposing Dual Self-Consistency Reinforcement Learning, we leverage the LaTeX toolchain for verifiable render-and-compare feedback. Experiments confirm our approach improves compilability and fidelity, establishing a foundation for visually-grounded synthesis.

\begin{acks}
This work is supported by Shanghai Artificial Intelligence Laboratory and the NSFC 62402426, the Soft Science Research Program of Zhejiang Province NO. 2026C35025, the Key R\&D Projects in Zhejiang Province (No. 2025C02156, 2025C01128), the Zhejiang NSF 2024C01106, Key Laboratory of Intelligent Processing Technology for Digital Music (Zhejiang Conservatory of Music), Ministry of Culture and Tourism NO. 202312025.
\end{acks}

\bibliographystyle{ACM-Reference-Format}
\balance
\bibliography{sample-base}

\end{document}